\documentclass[letterpaper]{article}
\usepackage{aaai2026}  
\usepackage{times}  
\usepackage{helvet}  
\usepackage{courier}  
\usepackage[hyphens]{url}  
\usepackage{graphicx} 
\usepackage{natbib}  
\usepackage{caption} 
\usepackage{algorithm}
\usepackage{multirow}
\usepackage{amsfonts} 
\usepackage{amsmath} 
\usepackage{xcolor}
\usepackage{algpseudocode}

\usepackage{tcolorbox}

\usepackage{newfloat}
\usepackage{listings}
\DeclareCaptionStyle{ruled}{labelfont=normalfont,labelsep=colon,strut=off} 
\floatstyle{ruled}
\newfloat{listing}{tb}{lst}{}
\floatname{listing}{Listing}

\title{Surgical AI Copilot: Energy-Based Fourier Gradient Low-Rank Adaptation for Surgical LLM Agent Reasoning and Planning}
\author{
    Jiayuan Huang\textsuperscript{\rm 1, \rm 2, \rm 6},
    Runlong He\textsuperscript{\rm 1, \rm 2},
    Danyal Zaman Khan\textsuperscript{\rm 1, \rm 3, \rm7},
    Evangelos B. Mazomenos\textsuperscript{\rm 1, \rm 2},
    Danail Stoyanov\textsuperscript{\rm 1, \rm 4},
    Hani Marcus\textsuperscript{\rm 1, \rm 3},
    Linzhe Jiang\textsuperscript{\rm 1, \rm 2},
    Matthew J Clarkson\textsuperscript{\rm 1, \rm 2},
    Mobarak I. Hoque\textsuperscript{\rm 1, \rm 2, \rm5}
}
\affiliations{
    \textsuperscript{\rm 1}UCL Hawkes Institute, University College London, UK\\
    \textsuperscript{\rm 2}Dept of Medical Physics \& Biomedical Engineering, University College London, UK\\
    \textsuperscript{\rm 3}Dept of Neurosurgery, National Hospital for Neurology and Neurosurgery, UK\\
    \textsuperscript{\rm 4}Dept of Computer Science, University College London, UK\\
    \textsuperscript{\rm 5}Division of Informatics, Imaging and Data Science, The University of Manchester, UK\\
    \textsuperscript{\rm 6}Visual Understanding Research Group, Dept of Informatics, King's College London, UK\\
    \textsuperscript{\rm 7}Institute of Neurology, University College London, UK\\
    jiayuan.1.huang@kcl.ac.uk, runlong.he.23@ucl.ac.uk, mobarak.hoque@manchester.ac.uk
}

\begin{document}

\maketitle

\begin{abstract}
Image-guided surgery demands adaptive, real-time decision support, yet static AI models struggle with structured task planning and providing interactive guidance. Large language models (LLMs)-powered agents offer a promising solution by enabling dynamic task planning and predictive decision support. Despite recent advances, the absence of surgical agent datasets and robust parameter-efficient fine-tuning techniques limits the development of LLM agents capable of complex intraoperative reasoning. In this paper, we introduce Surgical AI Copilot, an LLM agent for image-guided pituitary surgery, capable of conversation, planning, and task execution in response to queries involving tasks such as MRI tumor segmentation, endoscope anatomy segmentation, overlaying preoperative imaging with intraoperative views, instrument tracking, and surgical visual question answering (VQA). To enable structured agent planning, we develop the PitAgent dataset, a surgical context-aware planning dataset covering surgical tasks like workflow analysis, instrument localization, anatomical segmentation, and query-based reasoning. Additionally, we propose DEFT-GaLore, a Deterministic Energy-based Fourier Transform (DEFT) gradient projection technique for efficient low-rank adaptation of recent LLMs (e.g., LLaMA 3.2, Qwen 2.5), enabling their use as surgical agent planners.  We extensively validate our agent's performance and the proposed adaptation technique against other state-of-the-art low-rank adaptation methods on agent planning and prompt generation tasks, including a zero-shot surgical VQA benchmark, demonstrating the significant potential for truly efficient and scalable surgical LLM agents in real-time operative settings.
\end{abstract}
\begin{links}
    \link{Code}{https://github.com/mobarakol/SurgicalAICopilot}
\end{links}

\section{Introduction}

The integration of AI into image-guided robotic and laparoscopic surgery has the potential to revolutionize minimally invasive procedures by providing adaptive, interactive, and real-time decision support~\cite{maier2022surgical}. However, traditional AI models, often static and lacking the ability to perform structured task planning and surgical workflow reasoning, fall short~\cite{gao2024empowering}, making it challenging to coordinate tasks involving preoperative imaging, intraoperative video, tracking, and navigation~\cite{chen2024vs}. The challenge is particularly critical in endonasal pituitary surgery, where a narrow surgical corridor and proximity to critical neurovascular structures require highly context-aware support~\cite{khan2023current}. Modern large language models (LLMs) and vision-language models (VLMs), however, offer a promising avenue for addressing this challenge by enabling AI-driven systems that can interpret surgical scenes, respond to natural language queries, and dynamically plan actions based on the evolving intraoperative context~\cite{moghani2024sufia}. Building upon these advancements, there is a critical need for a specialized AI Agent that seamlessly integrates surgical workflow understanding, dynamic task planning, and interactive decision support. 

Currently, several AI Agents have been developed for various application, such as MedAgents~\cite{tang2023medagents} designed to mimic clinical roles, SurgRAW designed to coordinate multiple vision‑language agents using chain‑of‑thought reasoning in robotic-assisted surgery, and LlaVa-Surge designed to understand and answer open-ended questions about surgical videos.
However, these existing agents either rely solely on zero-shot inference or exhaustive fine-tuning, and critically lack specialized datasets designed explicitly for surgical agent planning and workflow reasoning. 
Without agent-planning datasets,surgical AI agents lack comprehensive domain knowledge, which leads to hallucinations. It makes them struggle to coordinate the complex, dynamic tasks of surgery, fail to systematically capture the structured sequence of surgical workflows, and have difficulty maintaining long-term context across surgical stages, ultimately limiting their real-time applicability and clinical reliability.

Parameter-Efficient Fine-Tuning (PEFT) is a set of techniques used to adapt large pre-trained models to new tasks without updating all of the model's parameters. Although current PEFT methods such as LoRA~\cite{hu2022lora} and DoRA~\cite{liu2024dora} can significantly reduce trainable parameters while preserving performance, they often struggle to adapt to highly dynamic and context-sensitive surgical environments and may fail to fully capture the intricate relationships between multimodal surgical context. Although methods like GaLore~\cite{zhao2024galore} and DCT-GaLore~\cite{modoranu2025svd} can enable memory-efficient optimization via gradient low-rank projections, they are not computationally efficient, and hard to capture time-varying dominant optimization direction in surgical scenarios.

In this paper, we address these limitations by introducing Surgical AI Copilot, the first surgical LLM agent planner for image-guided pituitary surgery, which is fine-tuned with a new DEFT-GaLore adaptation technique on our proposed PitAgent dataset to enable interactive planning, conversation, and action within this specific surgical context. Following the \textit{Planner–Worker} agent framework~\cite{xu2023rewoo, zeng2024agenttuning}, our copilot explicitly separates reasoning and execution into a \textit{Planner}, responsible for query decomposition and sub-task planning, and a \textit{Worker} that invokes specialized surgical models to perform the corresponding actions.
Our contributions are summarized as follows:

\begin{itemize}
    \item We introduce PitAgent dataset, the first surgical context-aware dataset designed for task planning in endonasal pituitary surgery, encompassing segmentation, overlaying, instrument localization, tracking, phase identification, and transsphenoidal-specific surgical activity recognition, providing structured workflow information to enhance surgical decision-making.
    
    \item We propose DEFT-GaLore, a novel deterministic energy-based Fourier transform gradient projection method for low-rank adaptation, enabling efficient fine-tuning of LLaMA 3.2~\cite{llama32vision} and Qwen 2.5~\cite{zhu2025qwen} to enhance real-time surgical decision support and task-specific adaptation.

    \item We evaluate our LLM-based agent’s task planning, prompt generation and execution capabilities, benchmarking its adaptation performance against existing techniques, and validate its surgical VQA outputs on a public pituitary dataset, demonstrating semantically meaningful responses and superior real-world applicability.
\end{itemize}

\section{Related Work}
\subsection{LLM based Agent in Medical Applications} 

The LLM-Agent represents a transformative advancement in surgical AI, capable of planning, memory retention, and real-time decision-making, making it an ideal interactive and collaborative copilot for surgeons~\cite{gao2025multimodal, kelly2024visiongpt}. Trained or fine-tuned on surgery-specific, large-scale datasets, the agent has the potential to plan sequences of tasks, communicate with other AI models or tools, and deliver evidence based support and visual assistance such as real time instrument-anatomy segmentation, tool tracking, diagnostic image segmentation, overlaying and visual question answering (VQA) onto intraoperative endoscopic views. Several clinical LLM-agents have been proposed and developed for various applications. For example, the AgentClinic benchmark~\cite{dutta2024adaptive}, an automated diagnostic agent architecture for use in simulated clinical environments, where the agent dynamically corrects misdiagnoses and interacts with simulated patients to improve diagnostic accuracy. MedAide~\cite{wei2024medaide}, a system uses multi-agent collaboration to provide integrated diagnostics and decision support, and performs well on multiple healthcare benchmarks.
Additionally, efforts to develop surgical AI agents include SurgBox~\cite{wu2024surgbox}, which simulates surgical workflows, and VS-Assistant~\cite{chen2024vs}, which provides on-demand visual support in surgery. However, these agents either rely primarily on retrieval-augmented generation (RAG)~\cite{gao2023retrieval} or rely solely on zero-shot inference or utilize simplistic classification-based call functions, neglecting comprehensive fine-tuning of LLMs for adaptive decision-making in surgical environments.

\begin{figure*}[ht]
    \centering
    \includegraphics[width=0.95\textwidth]{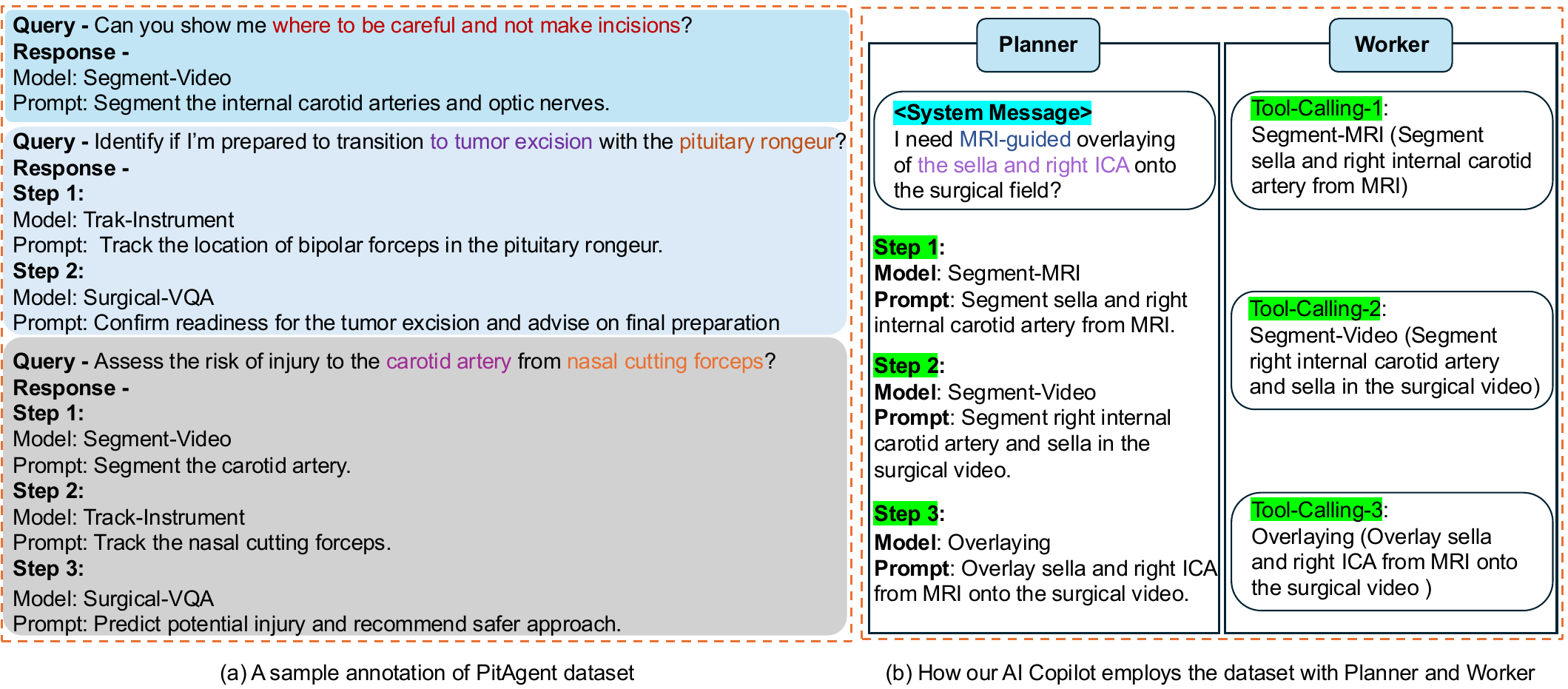}
    \caption{(a) denotes Query and Response samples of PitAgent dataset. (b) denotes how the \textit{Planner-Worker} architecture employs the dataset to enable multi-step decision-making and real-time guidance for surgical scenarios.}
    \label{fig:agent_dataset}
\end{figure*}

\subsection{PEFT LLM Finetuning}
 PEFT techniques can be applied to adapt LLMs efficiently to task-specific smaller datasets. There are currently two main categories of PEFT techniques designed to facilitate LLM adaptation. The first category comprises adaptation methods based on learning parameters, such as LoRA~\cite{hu2022lora}, MoRA~\cite{jiang2024mora}, and DoRA~\cite{liu2024dora}. The second category includes gradient low-rank projection approaches, such as GaLore~\cite{zhao2024galore}, GaLore 2~\cite{su2025galore} and DCT-GaLore~\cite{modoranu2025svd}. While these techniques have demonstrated strong performance in reducing the number of trainable parameters, they also present certain limitations. Learning parameter-based adaptation methods may struggle to generalize to out-of-distribution tasks, particularly when the injection of additional parameters is suboptimally placed, potentially leading to degraded performance~\cite{zhang2025parameter, han2024parameter, chen2023parameter}. On the other hand, gradient low-rank projection approaches, though efficient in terms of memory and computation, can introduce optimization instability when navigating complex loss landscapes~\cite{zhao2024galore}. This is especially true in high-dimensional gradient spaces, where the choice of projection direction may constrain the model's learning capacity. In addition, the lack of surgery-specific agent datasets that encompass planning, interaction, and action within the surgical context further limits the development of truly adaptive and interactive AI copilots for surgery.

\section{Methodology}
\subsection{Preliminaries: Gradient Low-Rank Projection}

GaLore~\cite{zhao2024galore} is a technique for optimizing neural network training, aiming to reduce memory usage by lowering the rank of gradient matrices. For the gradient matrix \({G}_t \in \mathbb{R}^{m \times n}\) at iteration \(t\) , GaLore applies singular value decomposition (SVD) with rank $k$ to obtain projection matrices \(P_t \in \mathbb{R}^{m \times r}\) and \(Q_t \in \mathbb{R}^{n \times r}\) as follows:

\begin{equation}
    \scalebox{0.96}{$
     G_t = USV^\top \approx \sum_{i=1}^{r} s_i u_i v_i^\top 
    $}
\end{equation}
where P$_t = [u_1, u_2, \dots, u_k], \ Q_t = [v_1, v_2, \dots, v_k]$. Then, the gradient update using the projected matrices is formulated as:
\begin{equation}
    \label{galore-w}
    \scalebox{0.96}{$
    W_T = W_0 + \eta \sum_{t=0}^{T-1} \left[ P_{t}\,\rho_t\Bigl(P_t^\top\, G_t\, Q_t\Bigr)\,Q_t^\top \right]
    $}
\end{equation}

\noindent Where \(\rho_t\)  is an entry-wise stateful gradient regularizer (e.g., Adam), \({P}_t^\top{G}_t{Q}_t\) is the low-rank matrix projected into the \(\mathbb{R}^{r \times r}\)space and \(\eta\) is the learning rate. While GaLore significantly reduces memory consumption by maintaining low-rank gradient statistics, the primary limitations are the computational expense of performing SVD.

GaLore 2~\cite{su2025galore} improves the SVD to a Fast Randomized SVD, but it is still essentially an SVD, although it reduces the training time, but it still essentially uses a high computational complexity SVD, and the primary limitation is not solved.

DCT-GaLore~\cite{modoranu2025svd} replaces SVD with DCT to construct an orthogonal basis and alleviates optimization information loss via a dynamic column selection strategy that selects basis vectors based on gradient similarity. However, since it samples only a few columns from a fixed, gradient-independent cosine basis containing only real components, phase information in the frequency domain is lost, reducing projection expressiveness and optimization efficiency.

\subsection{Proposed Method: Surgical AI Copilot}

\begin{figure*}[!ht]
    \centering
    \includegraphics[width=0.80\textwidth]{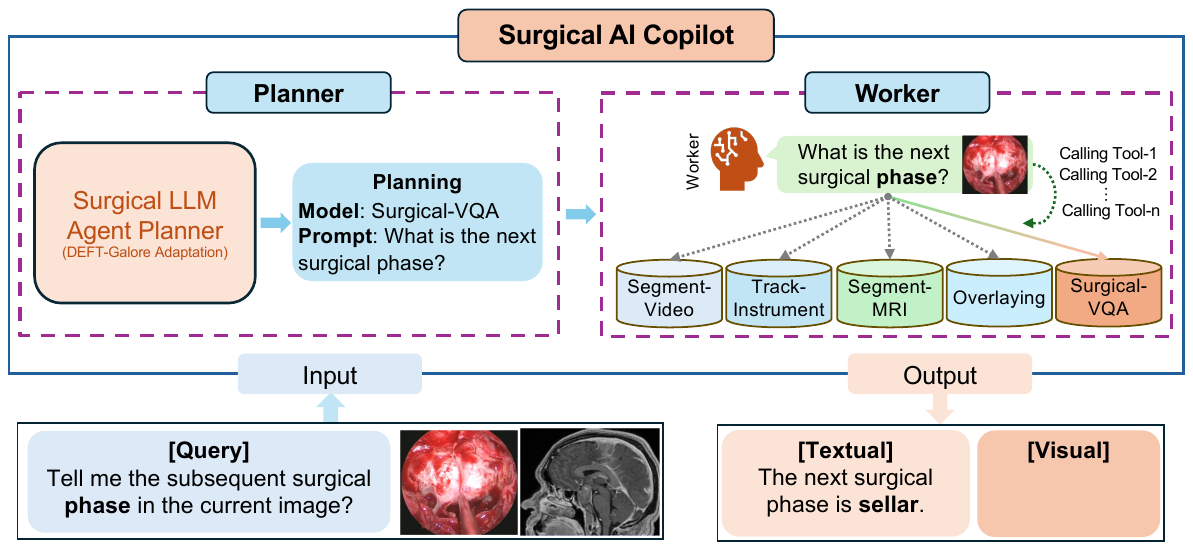}
    \caption{The \textit{Planner-Worker} architecture of our Surgical LLM Agent. The \textit{Planner} analyzes the surgeon's query and generates planning prompts. Based on these prompts, the \textit{Worker} dynamically calls multiple promptable AI models such as segmentation, tracking, and MRI registration to collaboratively answer surgical queries.}
    \label{fig:model_archi}
\end{figure*}

We design the Surgical LLM Copilot by creating the PitAgent dataset and DEFT-GaLore to adapt open-source LLMs (LLaMA 3.2 and Qwen 2.5) into a surgical AI agent. This agent enables task planning and prompt generation for promptable AI models supporting surgical tasks like VQA, video segmentation, MRI segmentation, overlaying, and instrument tracking. As shown in Figure~\ref{fig:model_archi}, our copilot integrates two core components: the \textit{Planner} and the \textit{Worker}. Together, they form an adaptive, interactive framework for pituitary surgery. The \textit{Planner} acts as the cognitive center, dynamically interpreting surgeon queries, decomposing them into structured sub-tasks, and generating plans with model-specific prompts based on real-time context. Once tasks are identified and sequenced, the \textit{Planner} delegates execution to the Worker. The \textit{Worker} leverages specialized promptable visual and multimodal models to execute these tasks using multimodal inputs. After execution, the agent returns visual or textual results back to the user. This continuous interplay between the \textit{Planner}'s context understanding and the \textit{Worker}'s multimodal processing ensures precise, real-time, and contextually relevant surgical assistance.

\begin{table}[!ht]
\centering
\caption{Key content and number of unique queries of each category in PitAgent dataset. Q. denotes unique quantity of queries.}
\label{table:keyContent}
\scalebox{0.72}{
\begin{tabular}{c|c|l}
\hline
\textbf{Category} & \textbf{Q.} & \multicolumn{1}{c}{\textbf{Key Content}} \\ \hline
\begin{tabular}[c]{@{}c@{}}Surgical-VQA\end{tabular} & 52 & \begin{tabular}[c]{@{}l@{}}Surgical step, Surgical phase, Next surgical phase,\\ Next surgical step, Number of instruments,\\ Instrument location, Surgical activity, \\Instrument recognition, Remaining time\end{tabular} \\ \hline
\begin{tabular}[c]{@{}c@{}}Segment-Video\end{tabular} & 27 & \begin{tabular}[c]{@{}l@{}}Sella, Clival Recess, Left Carotid,\\ Right Carotid, Left Optic Protuberance,\\ Right Optic Protuberance, Carotid,\\ Optic Protuberance, All Anatomical \\(Neurovascular) Structures\end{tabular} \\ \hline
\begin{tabular}[c]{@{}c@{}}Segment-MRI\end{tabular} & 15 & \begin{tabular}[c]{@{}l@{}}Pituitary Adenoma, Pituitary Tumor,\\ Internal Carotid Artery (ICA)\end{tabular} \\ \hline
\begin{tabular}[c]{@{}c@{}}Track-Instrument\end{tabular} & 27 & \begin{tabular}[c]{@{}l@{}}Bipolar Forceps, Cottle Elevator, Cup Forceps,\\ Dural Scissors, Freer Elevator, Hemostatic Foam,\\Irrigation Syringe, Kerrison Rongeur, \\ Micro-Doppler Probe, Nasal Cutting Forceps,\\ Pituitary Rongeur, Retractable Knife,\\ Ring Curette, Spatula Dissector, Stealth Pointer,\\ Suction Cannula, Surgical Drill, Tissue Glue\end{tabular} \\ \hline
\begin{tabular}[c]{@{}c@{}}Multitask-1\\ (2 models)\end{tabular} & 90 & \begin{tabular}[c]{@{}l@{}}Segment-Video→Segment-MRI, \\Segment-Video→Track-Instrument,\\ Segment-Video→Surgical-VQA,\\ Track-Instrument→Surgical-VQA,\\ Overlaying→Surgical-VQA,\\ Surgical-VQA →Surgical-VQA\end{tabular} \\ \hline
\begin{tabular}[c]{@{}c@{}}Multitask-2\\ (3 models)\end{tabular} & 50 & \begin{tabular}[c]{@{}l@{}}Segment-Video→Track-Instrument→Surgical-VQA,\\ Surgical-VQA→Surgical-VQA→Surgical-VQA,\\ Model1→Model2→Overlaying\end{tabular} \\ \hline
\end{tabular}
}
\end{table}

\subsubsection{PitAgent Dataset}
\label{subsubsec:pitagent}
Currently, there is a lack of specialized datasets in the field of endonasal pituitary surgery that are explicitly designed for surgical agent planning and workflow reasoning. To address this gap, we introduce PitAgent, a surgical-domain dataset designed to facilitate fine-tuning of our surgical LLM copilot planner for performing reasoning over queries, available inputs and models to generate plans and corresponding prompts.

During the dataset construction process, and with the support of our clinical collaborators, we first identified queries that closely reflect those a surgeon would naturally pose in real surgical scenarios across 6 query-response categories. As shown in Table~\ref{table:keyContent},  Four categories represent the agent planning with single-model tasks, and the remaining two categories (Multitask-1 and Multitask-2) involve coordinating multiple base models. 
Based on the complexity of each query, we annotated the agent planning sequences with the appropriate model selection along with the prompt required to invoke the corresponding model, which serves as the response to the query. The planning annotations are aligning publicly available pituitary dataset of endoscopic videos, tasks of video segmentation~\cite{mao2024pitsurgrt}, MRI segmentation~\cite{zhao2024transferring}, tracking~\cite{islam2020ap}, overlaying~\cite{enkaoua2023image}, and Surgical-VQA~\cite{he2024pitvqa}. Some examples of the annotated samples are illustrated in Figure~\ref{fig:agent_dataset}(a), it shows how our LLM copilot employs the dataset with \textit{Planner} and \textit{Worker}. Subsequently, all query-response pairs were reviewed and validated by clinical collaborators. The final number of unique query-response pairs per category is summarized in Table~\ref{table:keyContent}. Among these categories, the Multitask-1 category contains the largest number of pairs (90), while Segment-MRI includes the fewest (15). The unique question numbers of Surgical-VQA, Segment-Video, Track-Instrument, and Multitask-2 are 52, 27, 27, and 50 respectively. Each finalized unique query was further rephrased 50 times, resulting in a total of 13,050 query-response pairs. Table~\ref{table:keyContent} also details the scope of each task: Surgical-VQA covers 8 surgical concepts, Segment-Video handles 6 anatomical structures, Segment-MRI focuses on 3 anatomical structures, and Track-Instrument includes 18 instrument types. multitasks includes two or three combinations of promptable models. Finally, we split the query-response pairs into training and testing sets in an 8:2 ratio.
\begin{figure}[!ht]
    \centering
    \includegraphics[width=0.48\textwidth]{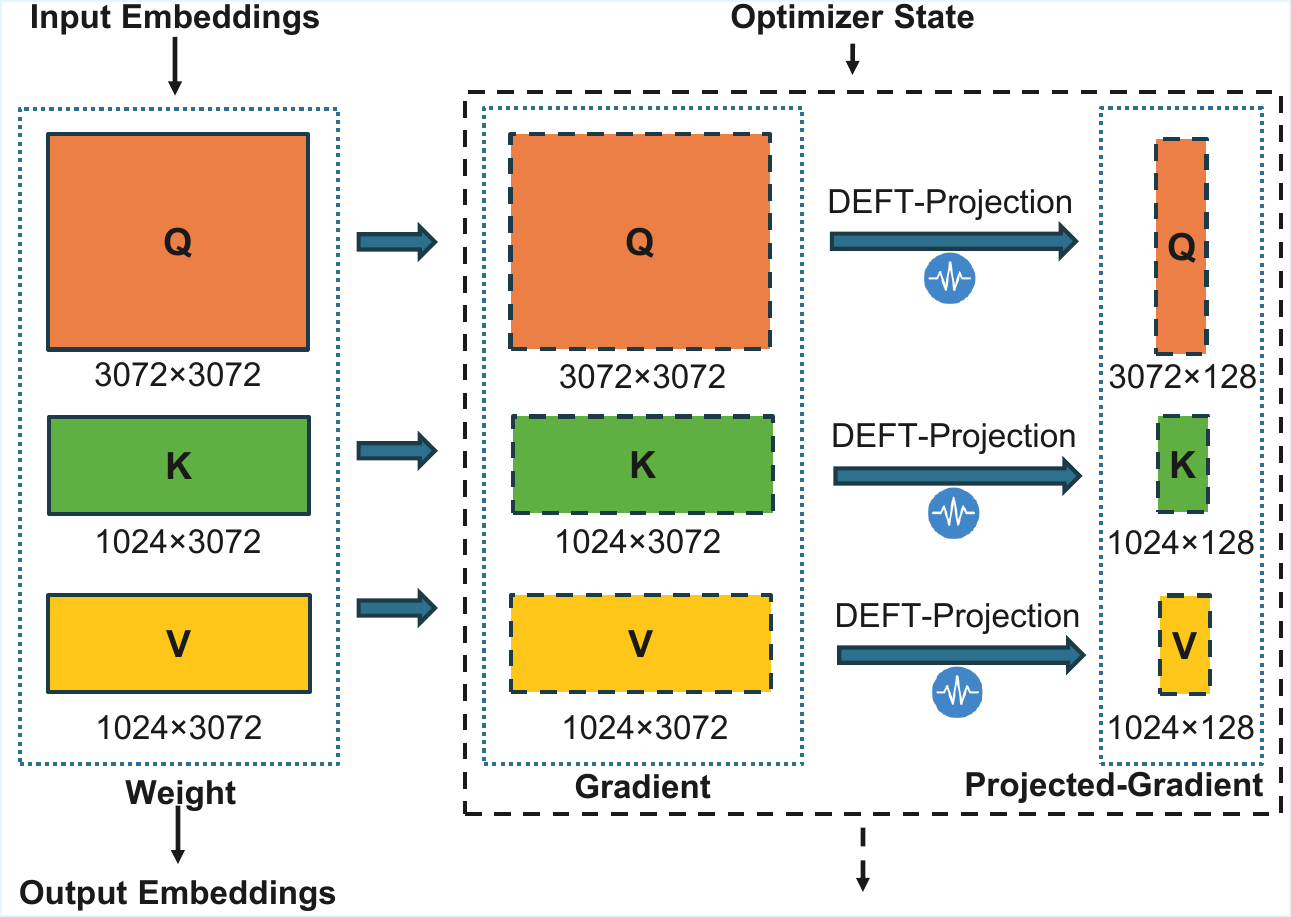}
    \caption{Example of constructing DEFT-GaLore projection gradient for LLaMA 3.2 model.}
    \label{fig:projector}
\end{figure}

\subsubsection{Deterministic Energy-based Fourier Transform Gradient Low-rank Projection}
To overcome the bottleneck caused by aforementioned gradient low-rank projection techniques, we introduce Deterministic Energy-Based Fourier Transform GaLore (DEFT-GaLore), a novel gradient low-rank projection method that replaces SVD with an efficient frequency-domain analysis. Energy is defined as the sum of squared magnitudes, which measures the informational content of a matrix and is numerically equivalent to the square of its Frobenius norm~\cite{reza2019class}.
The proposed method rests on two theoretical pillars: (i) Projection matrix construction based on Frobenius norms minimises information loss due to projection~\cite{modoranu2025svd}, and (ii) Parseval’s theorem guarantees that the energy computed in the frequency domain equals that in the time (or spatial) domain~\cite{fein2025fft}.

DEFT-GaLore is proposed to compress full-rank gradients into a low-rank subspace. To this end, the method first transforms into the frequency domain using a Fast Fourier Transform (FFT), computes the spectral energy of the matrix, and then extracts a sketch matrix corresponding to the dominant frequencies. Finally, a real-valued orthogonal projection matrix is constructed from this sketch. Specifically, given an input matrix $G_t \in \mathbb{R}^{m \times n}$ at step $t$ and target rank $k$, the DEFT-GaLore algorithm computes a low-rank orthogonal projection matrix $M_t \in \mathbb{R}^{m \times k}$ through the following steps:

\begin{enumerate}
    \item \textbf{Fast Fourier Transform (FFT)}: Apply FFT along the column dimension to $G_t$:
    \begin{equation} \label{eq:fft}
        G_{t,f} = \mathcal{F}(G_t) \in \mathbb{R}^{m \times n}
    \end{equation}
    where $G_t$ is real-valued input matrix at step $t$ (time/spatial domain); $G_{t,f}$ is complex frequency-domain representation after FFT; $\mathcal{F}(\cdot)$ is the Fast Fourier Transform applied column-wise.
    
    \item \textbf{Energy Spectrum Calculation}: Compute the energy for each frequency component:
    \begin{equation} \label{eq:energy}
        s_j = \sum_{i=1}^{m} \left| G_{t,f}(i,j) \right|^2, \quad j = 1, 2, \dots, n
    \end{equation}
    where $s_j$ is energy value at frequency index $j$; $G_{t,f}(i,j)$is complex element at row $i$, column $j$;
    $|\cdot|$is Complex modulus operator.
    
    \item \textbf{Frequency Selection}: Identify the top-$k$ frequency indices with maximum energy:
    \begin{equation} \label{eq:selection}
        \mathcal{I} = \{ j_1, j_2, \dots, j_k \} \subseteq \{ 1, 2, \dots, n \}
    \end{equation}
    where $\mathcal{I}$ is the index set sorted in ascending order.
    
    \item \textbf{Sketch Matrix Construction}: Extract columns corresponding to $\mathcal{I}$:
    \begin{equation} \label{eq:sketch}
        Y_c = G_{t,f}(: , \mathcal{I}) \in \mathbb{C}^{m \times k}
    \end{equation}
    where ${Y}_c$ is the complex sketch matrix.
    
    \item \textbf{Complex QR Decomposition}: Perform QR decomposition on the sketch matrix:
    \begin{equation} \label{eq:complex_qr}
        Q_c, R_c = \text{QR}(Y_c), \quad Q_c \in \mathbb{C}^{m \times k}, \  R_c \in \mathbb{C}^{k \times k}
    \end{equation}
    where ${Q}_c$ is the complex orthogonal basis matrix.
    
    \item \textbf{Real-Imaginary Separation}: Decompose ${Q}_c$ into real components:
    \begin{equation} \label{eq:realimag}
        Y_r = \text{reshape}\left( \text{RealImag}(Q_c) \right) \in \mathbb{R}^{m \times 2k}
    \end{equation}
    where $\text{RealImag}({Q}_c)$ splits each complex element $q = a + bi$ into $(a, b)$; $\text{reshape}$ converts the $\mathbb{C}^{m \times k}$ matrix to $\mathbb{R}^{m \times 2k}$
    
    \item \textbf{Real QR Decomposition}: Perform QR decomposition on the extended real matrix:
    \begin{equation} \label{eq:real_qr}
        Q_r, R_r = \text{QR}(Y_r), \quad Q_r \in \mathbb{R}^{m \times 2k}, \  R_r \in \mathbb{R}^{2k \times 2k}
    \end{equation}
    
    \item \textbf{Final Projection Matrix}: Extract the first $k$ columns:
    \begin{equation} \label{eq:final_q}
        M_t = Q_r(: , 1:k) \in \mathbb{R}^{m \times k}
    \end{equation}
\end{enumerate}

\noindent The complete DEFT-GaLore projection is summarized as:
\begin{equation} \label{eq:full}
    \scalebox{0.95}{$M_t = \text{QR}\left[ \text{reshape}\left( \text{RealImag}\left( \text{QR}\left({G}_{t,f}(: , \mathcal{I}) \right) \right) \right) \right]_{(: , 1:k)}$}
\end{equation}

\noindent A similar process can be applied to the transpose of the gradient matrix to obtain the right projection matrix $N_t \in \mathbb{R}^{n \times k}$.
Finally, given the pretrained weight \({W}_0 \in \mathbb{R}^{m \times n}\) the gradient update using the projected matrices for our DEFT-GaLore can be derived as:
\begin{equation}
    \label{deft-galore-w}
    \scalebox{0.96}{$
    W_T = W_0 + \eta \sum_{t=0}^{T-1} \left[ M_{t}\,\rho_t\!\Bigl(M_{t}^{\top}\, G_t\,N_{t}\Bigr)\,N_{t}^\top \right]
    $}
\end{equation}

DCT-GaLore is constrained by its reliance on a fixed global orthogonal basis composed exclusively of real cosine functions with even symmetry. When this basis is truncated to a low rank, it can only adjust the amplitude, while the phase information, implicitly encoded in higher-frequency components, is lost during truncation.
In comparison, our approach enables both amplitude and phase information to be preserved in the low-rank projection, and it selects the columns with the highest energy, allowing for more accurate maintenance of gradient directionality in low-rank settings.

As shown in Figure~\ref{fig:projector}, we integrated DEFT-GaLore into every transformer block of the LLaMA 3.2 models, specifically applying it to the query projection (Q), key projection (K), value projection (V), and output projection (O) layers, while maintaining a projection rank of 128 for a fair comparison with the original GaLore.

\begin{table*}[!ht]
\centering
\caption{Comparison of the performance of various PEFT methods in terms of Prompt Generation, Planning, and Training Overhead. Results cover PEFT methods for the zero-shot LLMs and two PEFT-fine-tune models, with bold being the best in the same segment.}
\label{tab:at}
\resizebox{0.95\textwidth}{!}{
\begin{tabular}{cccccccc}
\hline
\multicolumn{1}{c|}{\multirow{2}{*}{Method}} & \multicolumn{4}{c|}{\textbf{Prompt Generation}} & \multicolumn{2}{c|}{\textbf{Planning}} & \textbf{Training Overhead} \\ \cline{2-8} 
\multicolumn{1}{c|}{} & \multicolumn{1}{c|}{\begin{tabular}[c]{@{}c@{}}BLEU-3\\ (\%)\end{tabular}} & \multicolumn{1}{c|}{\begin{tabular}[c]{@{}c@{}}BLEU-4\\ (\%)\end{tabular}} & \multicolumn{1}{c|}{\begin{tabular}[c]{@{}c@{}}ROUGE-L\\ (\%)\end{tabular}} & \multicolumn{1}{c|}{\begin{tabular}[c]{@{}c@{}}METEOR\\ (\%)\end{tabular}} & \multicolumn{1}{c|}{\begin{tabular}[c]{@{}c@{}}F1\\ (\%)\end{tabular}} & \multicolumn{1}{c|}{\begin{tabular}[c]{@{}c@{}}ACC\\ (\%)\end{tabular}} & \begin{tabular}[c]{@{}c@{}}Time\\ (second)\end{tabular} \\ \hline
\multicolumn{8}{c}{\textbf{Zero-shot}} \\ \hline
\multicolumn{1}{c|}{LLaMA 3.2~\cite{llama32vision}} & \multicolumn{1}{c|}{10.88} & \multicolumn{1}{c|}{8.21} & \multicolumn{1}{c|}{44.89} & \multicolumn{1}{c|}{52.05} & \multicolumn{1}{c|}{48.17} & \multicolumn{1}{c|}{32.5} & --- \\
\multicolumn{1}{c|}{Qwen 2.5~\cite{zhu2025qwen}} & \multicolumn{1}{c|}{3.41} & \multicolumn{1}{c|}{2.19} & \multicolumn{1}{c|}{21.25} & \multicolumn{1}{c|}{36.64} & \multicolumn{1}{c|}{45.33} & \multicolumn{1}{c|}{31.65} & --- \\
\multicolumn{1}{c|}{Gemma 2~\cite{team2024gemma}} & \multicolumn{1}{c|}{3.03} & \multicolumn{1}{c|}{2.21} & \multicolumn{1}{c|}{12.25} & \multicolumn{1}{c|}{27.55} & \multicolumn{1}{c|}{26.67} & \multicolumn{1}{c|}{18.42} & --- \\
\multicolumn{1}{c|}{Deepseek llm~\cite{bi2024deepseek}} & \multicolumn{1}{c|}{16.30} & \multicolumn{1}{c|}{12.40} & \multicolumn{1}{c|}{43.64} & \multicolumn{1}{c|}{49.49} & \multicolumn{1}{c|}{42.67} & \multicolumn{1}{c|}{30.22} & --- \\ \hline
\multicolumn{8}{c}{\textbf{LLaMA-3.2-3B PEFT}} \\ \hline
\multicolumn{1}{c|}{LoRA~\cite{hu2022lora}} & \multicolumn{1}{c|}{29.47} & \multicolumn{1}{c|}{25.14} & \multicolumn{1}{c|}{63.78} & \multicolumn{1}{c|}{70.02} & \multicolumn{1}{c|}{70.5} & \multicolumn{1}{c|}{61.46} & --- \\
\multicolumn{1}{c|}{DoRA~\cite{liu2024dora}} & \multicolumn{1}{c|}{51.57} & \multicolumn{1}{c|}{46.32} & \multicolumn{1}{c|}{71.76} & \multicolumn{1}{c|}{75.48} & \multicolumn{1}{c|}{77.17} & \multicolumn{1}{c|}{64.65} & --- \\
\multicolumn{1}{c|}{MoRA~\cite{jiang2024mora}} & \multicolumn{1}{c|}{33.52} & \multicolumn{1}{c|}{28.88} & \multicolumn{1}{c|}{65.12} & \multicolumn{1}{c|}{69.43} & \multicolumn{1}{c|}{69.33} & \multicolumn{1}{c|}{58.16} & --- \\
\multicolumn{1}{c|}{GaLore~\cite{zhao2024galore}} & \multicolumn{1}{c|}{74.11} & \multicolumn{1}{c|}{70.68} & \multicolumn{1}{c|}{87.81} & \multicolumn{1}{c|}{89.12} & \multicolumn{1}{c|}{97.67} & \multicolumn{1}{c|}{94.94} & 8762.43 \\
\multicolumn{1}{c|}{GaLore 2~\cite{su2025galore}} & \multicolumn{1}{c|}{73.18} & \multicolumn{1}{c|}{70.28} & \multicolumn{1}{c|}{88.38} & \multicolumn{1}{c|}{90.54} & \multicolumn{1}{c|}{98.17} & \multicolumn{1}{c|}{94.85} & 6891.79 \\
\multicolumn{1}{c|}{DCT-GaLore~\cite{modoranu2025svd}} & \multicolumn{1}{c|}{74.15} & \multicolumn{1}{c|}{71.21} & \multicolumn{1}{c|}{87.24} & \multicolumn{1}{c|}{89.96} & \multicolumn{1}{c|}{97.17} & \multicolumn{1}{c|}{92.76} & 3535.69 \\
\multicolumn{1}{c|}{DEFT-GaLore (ours)} & \multicolumn{1}{c|}{\textbf{74.27}} & \multicolumn{1}{c|}{\textbf{71.88}} & \multicolumn{1}{c|}{\textbf{89.87}} & \multicolumn{1}{c|}{\textbf{92.40}} & \multicolumn{1}{c|}{\textbf{99.00}} & \multicolumn{1}{c|}{\textbf{96.25}} & 3355.75 \\ \hline
\multicolumn{8}{c}{\textbf{Qwen-2.5-1.5B PEFT}} \\ \hline
\multicolumn{1}{c|}{LoRA~\cite{hu2022lora}} & \multicolumn{1}{c|}{12.48} & \multicolumn{1}{c|}{8.39} & \multicolumn{1}{c|}{35.97} & \multicolumn{1}{c|}{41.92} & \multicolumn{1}{c|}{25.83} & \multicolumn{1}{c|}{15.81} & --- \\
\multicolumn{1}{c|}{DoRA~\cite{liu2024dora}} & \multicolumn{1}{c|}{27.54} & \multicolumn{1}{c|}{22.87} & \multicolumn{1}{c|}{57.15} & \multicolumn{1}{c|}{67.32} & \multicolumn{1}{c|}{78.33} & \multicolumn{1}{c|}{69.86} & --- \\
\multicolumn{1}{c|}{MoRA~\cite{jiang2024mora}} & \multicolumn{1}{c|}{8.94} & \multicolumn{1}{c|}{5.47} & \multicolumn{1}{c|}{33.94} & \multicolumn{1}{c|}{40.85} & \multicolumn{1}{c|}{26.5} & \multicolumn{1}{c|}{16.3} & --- \\
\multicolumn{1}{c|}{GaLore~\cite{zhao2024galore}} & \multicolumn{1}{c|}{52.24} & \multicolumn{1}{c|}{47.68} & \multicolumn{1}{c|}{73.15} & \multicolumn{1}{c|}{77.09} & \multicolumn{1}{c|}{88.67} & \multicolumn{1}{c|}{79.07} & 2733.62 \\
\multicolumn{1}{c|}{GaLore 2~\cite{su2025galore}} & \multicolumn{1}{c|}{51.11} & \multicolumn{1}{c|}{46.29} & \multicolumn{1}{c|}{72.70} & \multicolumn{1}{c|}{76.42} & \multicolumn{1}{c|}{88.83} & \multicolumn{1}{c|}{75.15} & 2033.09 \\
\multicolumn{1}{c|}{DCT-GaLore~\cite{modoranu2025svd}} & \multicolumn{1}{c|}{44.81} & \multicolumn{1}{c|}{40.39} & \multicolumn{1}{c|}{67.68} & \multicolumn{1}{c|}{72.88} & \multicolumn{1}{c|}{85.17} & \multicolumn{1}{c|}{67.29} & 1775.41 \\
\multicolumn{1}{c|}{DEFT-GaLore (ours)} & \multicolumn{1}{c|}{\textbf{59.3}} & \multicolumn{1}{c|}{\textbf{55.53}} & \multicolumn{1}{c|}{\textbf{78.72}} & \multicolumn{1}{c|}{\textbf{84.1}} & \multicolumn{1}{c|}{\textbf{95.33}} & \multicolumn{1}{c|}{\textbf{91.96}} & 1764.88 \\ \hline
\end{tabular}
}
\end{table*}

\section{Experiments and Results}
\subsection{Dataset}

In addition to our PitAgent dataset, we also test our agent generated question prompts using a publicly available dataset of open-ended PitVQA~~\cite{he2025pitvqa++}. The dataset, derived from 25 videos of pituitary surgery~~\cite{he2024pitvqa}, consists of 59 unique questions spanning 6 key aspects, including surgical phases, steps, instruments, tool-tissue interaction, position and quantity.
To evaluate our agent-generated prompts, we retained the original validation sets~~\cite{he2025pitvqa++}, which include 24,767 frames and 182,720 question-answer pairs.

\subsection{Implementation Details}
The experiments are based on the LLaMA-3.2-3B-Instruct~\cite{llama32vision} and Qwen-2.5-1.5B-Instruct~\cite{zhu2025qwen} models provided by HuggingFace~\footnote{https://huggingface.co/meta-llama}. We first test the zero-shot performance of task planning and prompt generation abilities on above-mentioned models and two extra models, they are Gemma-2-2b~\cite{team2024gemma} and Deepseek-llm-7b-chat~\cite{bi2024deepseek}. Then We fine-tune the LLaMA and Qwen models with a cross-entropy loss and the AdamW optimizer, setting the learning rate and rank to \(3 \times 10^{-7}\) and 128 respectively. 
We compare our DEFT-GaLore with various parameter-efficient fine-tuning (PEFT) methods including LoRA~\cite{hu2022lora}, MoRA~\cite{jiang2024mora}, DoRA~\cite{liu2024dora}, GaLore~\cite{zhao2024galore}, GaLore2~\cite{su2025galore}, and DCT-GaLore~\cite{modoranu2025svd}. We also use the generated prompts from our agent as queries to the open-source PitVQA++ network~\cite{he2025pitvqa++} to further evaluate its performance on visual question answering tasks. All experiments are conducted within the PyTorch framework on an NVIDIA RTX A100\_80 GPU. More details such as \textit{System Message} used, \textit{external validation} and \textit{Pseudo-code} are included in the technical appendix. 
\begin{figure*}[!ht]
    \centering
    \includegraphics[width=\textwidth]{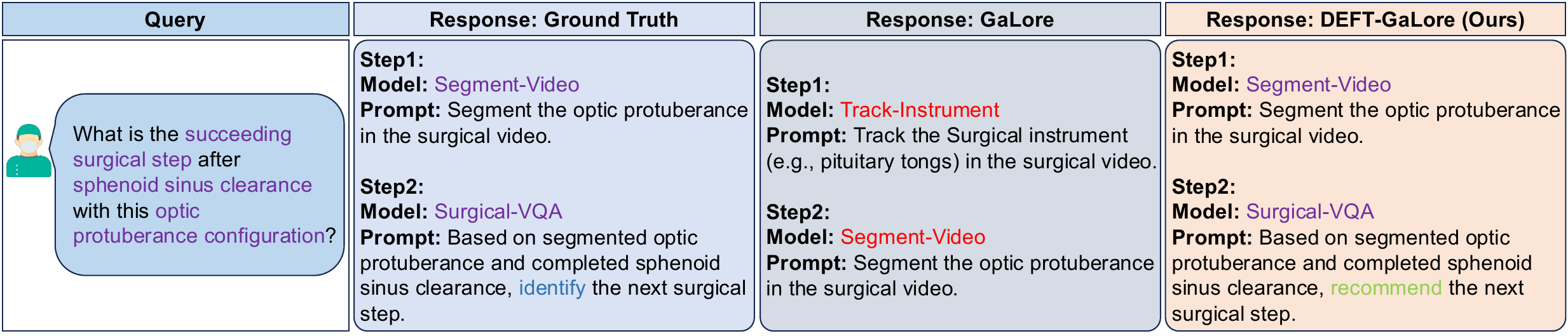}
    \caption{Qualitative results analysis of agent planning. The responses are generated by Ground Truth, GaLore fine-tuned \textit{Planner} and our DEFT-GaLore fine-tuned \textit{Planner}.}
    \label{fig:planning}
\end{figure*}

\subsection{Results}

\subsubsection{\textit{Planner}'s Performance}
The performance of the \textit{Planner} is evaluated from two aspects: task planning and prompt generation. We evaluate prompt generation quality using BLEU, ROUGE, and METEOR, and assess its task planning accuracy using accuracy (ACC) and F1 scores.
Table~\ref{tab:at} presents the results of performance of four zero-shot models. It also presents the performance of \textit{Planner} fine-tuned by our DEFT-GaLore compared with that fine-tuned by other PEFT methods. The zero-shot performance reveals that existing LLMs exhibit disappointing capabilities in task planning and prompt generation within the domain of pituitary surgery. With LLaMA as the agent \textit{Planner}, our method improved BLEU-4, METEOR, and F1 scores by 1.20\%, 3.28\%, and 1.33\% over the GaLore baseline on the PitAgent dataset. When using Qwen, the corresponding improvements reached 8.85\%, 7.01\%, and 12.89\%. Similar trends are observed in other metrics of ROUGE and ACC. Building upon this, our DEFT-GaLore achieves best‑in‑class results on all metrics across both PEFT model lines. It outperforms other state-of-the-art PEFT methods in both prompt generation and planning accuracy.
Figure \ref{fig:planning} visualizes the task‑planning results of GaLore and DEFT‑GaLore. Compared with the ground truth, the \textit{Planner} fine-tuned with GaLore misinterprets the query and selects the wrong model in both steps; the prompt it generates also differs greatly in meaning from the correct prompt. In contrast, DEFT‑GaLore not only performs task planning accurately but also produces a prompt whose structure closely mirrors the reference prompt and conveys the same semantics.

Table~\ref{tab:at} also presents the average time consumption of GaLore, DCT-GaLore and our DEFT-GaLore on LLaMA and Qwen models for fine-tuning one epoch. For LLaMA model, GaLore 2, DCT-GaLore and our DEFT-GaLore reduce the training time by 21.35\% 59.65\% and 61.7\% respectively compared to GaLore. For the Qwen model, the training time for both is reduced by 25.63\%, 35.04\% and 35.44\%, respectively.

\subsubsection{\textit{Worker}'s Performance}
\begin{figure}[!h]
    \centering
    \includegraphics[width=0.48\textwidth]{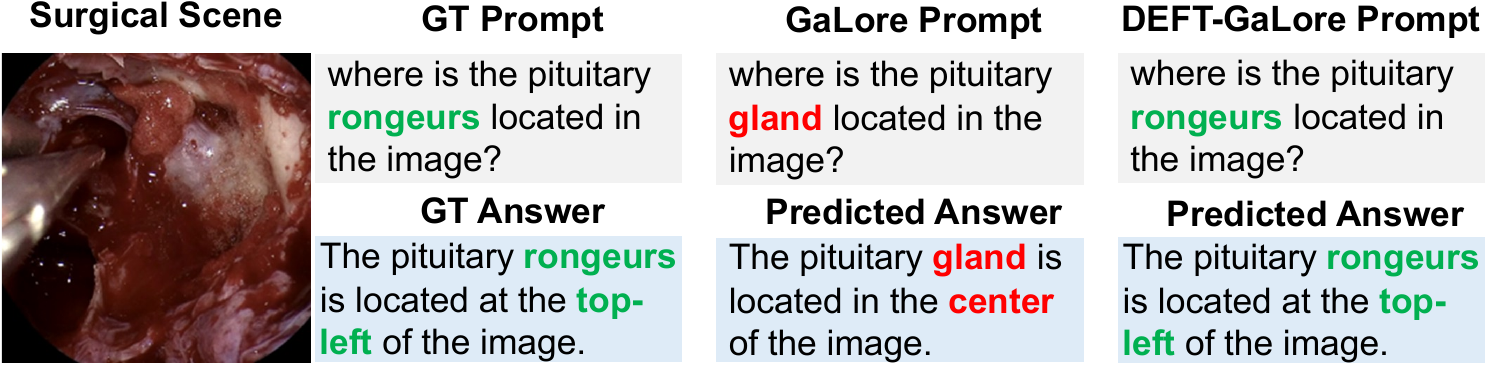}
    \caption{Qualitative result analysis of \textit{Worker} on \textit{Planner}-generated prompt vs GT prompt using publicly available zero-shot PitVQA++ model~\cite{he2025pitvqa++}.}
    \label{fig:quali_analysis}
\end{figure}

The performance of the \textit{Worker} is evaluated by assessing the downstream model's output when given prompts generated by the planner as input. Table~\ref{tab:vqaTest} compares \textit{Worker}'s performance using agent-generated prompts versus ground-truth prompts on a publicly available open-ended PitVQA dataset~\cite{he2025pitvqa++}. We evaluated our agent’s response quality in surgical VQA using two publicly available VLMs: LLaMA-3.2-Vision-11B~\cite{llama32vision} and PitVQA++~\cite{he2025pitvqa++}, where the performance metrics for \textit{Planner}-generated prompts are obtained from zero-shot PitVQA++ with its pretrained weights. The evaluation used standard metrics such as BLEU, ROUGE, and METEOR scores, measuring the impact of agent-generated prompts on model performance. While agent-generated prompts showed slightly lower performance than GT prompts (on average decreases of 19.44\% and 16.24\% in BLEU-4 and METEOR), our DEFT-GaLore outperformed other methods, improving BLEU-4 and METEOR scores by 3.42\% and 3.48\% over the GaLore baseline. Figure~\ref{fig:quali_analysis} shows qualitative results for surgical VQA. Prompt generation errors can affect answer accuracy, for instance, when the agent confused "rongeurs" with "gland", the downstream PitVQA++ model failed to locate the pituitary rongeurs.

\begin{table}[!ht]
\centering
\caption{\textit{Worker}'s performance with agent-generated vs. ground-truth prompts (zero-shot PitVQA++). Metrics: BLEU, ROUGE-L and METEOR.}
\label{tab:vqaTest}
\scalebox{0.68}{%
\begin{tabular}{cc|c|c|c|c}
\hline
\multicolumn{2}{c|}{\multirow{2}{*}{\textbf{Model}}} & \multirow{2}{*}{\textbf{\begin{tabular}[c]{@{}c@{}}B-3\\ (\%)\end{tabular}}} & \multirow{2}{*}{\textbf{\begin{tabular}[c]{@{}c@{}}B-4\\ (\%)\end{tabular}}} & \multirow{2}{*}{\textbf{\begin{tabular}[c]{@{}c@{}}R-L\\ (\%)\end{tabular}}} & \multirow{2}{*}{\textbf{\begin{tabular}[c]{@{}c@{}}MET.\\ (\%)\end{tabular}}} \\
\multicolumn{2}{c|}{} &  &  &  &  \\ \hline
\multicolumn{1}{c|}{\multirow{2}{*}{\begin{tabular}[c]{@{}c@{}}Ground Truth\\ Prompt\end{tabular}}} & PitVQA++~\cite{he2025pitvqa++} & 78.55 & 76.37 & 84.64 & 84.39 \\
\multicolumn{1}{c|}{} & LLaMA-Vision & 24.52 & 18.44 & 53.86 & 56.96 \\ \hline
\multicolumn{1}{c|}{\multirow{7}{*}{\begin{tabular}[c]{@{}c@{}}Agent-generated\\ Prompt\\ (LLaMA 3.2)\end{tabular}}} & LoRA~\cite{hu2022lora} & 36.39 & 31.74 & 46.41 & 41.90 \\
\multicolumn{1}{c|}{} & DoRA~\cite{liu2024dora} & 52.01 & 47.56 & 64.63 & 65.37 \\
\multicolumn{1}{c|}{} & MoRA~\cite{jiang2024mora} & 47.72 & 42.19 & 57.27 & 53.38 \\
\multicolumn{1}{c|}{} & GaLore~\cite{zhao2024galore} & 69.57 & 67.31 & 77.76 & 76.82 \\
\multicolumn{1}{c|}{} & GaLore 2~\cite{su2025galore} & 71.51 & 68.98 & 80.10 & 79.60 \\
\multicolumn{1}{c|}{} & DCT-GaLore (Mod. et al. 2025)  & 72.52 & 70.01 & 80.23 & 79.69 \\
\multicolumn{1}{c|}{} & \begin{tabular}[c]{@{}c@{}}DEFT-GaLore (ours)\end{tabular} & \textbf{73.31} & \textbf{70.73} & \textbf{80.77} & \textbf{80.3} \\ \hline
\end{tabular}
}
\end{table}

\subsubsection{Ablation Study}
Table~\ref{tab:rank} illustrates that rank-128 performs the best on LLaMA 3.2 model among all ranks tested. Both excessively low and high ranks slightly degrade the performance metrics. This indicates that the DEFT-GaLore method is sensitive to the rank hyperparameter: a rank that is too low may lead to underfitting due to the loss of optimization information, while a rank that is too high may introduce noise and cause overfitting.

\begin{table}[!ht]
\centering
\caption{Comparison of performance of LLaMA-based \textit{Planner} fine-tuned with DEFT-GaLore on different rank. The rank is set to 64, 128, and 256, respectively. }
\label{tab:rank}
\resizebox{0.45\textwidth}{!}{%
\begin{tabular}{c|cccc|cc}
\hline
\multirow{2}{*}{Rank} & \multicolumn{4}{c|}{Prompt Generation} & \multicolumn{2}{c}{Planning} \\ \cline{2-7} 
 & \multicolumn{1}{c|}{\begin{tabular}[c]{@{}c@{}}BLEU-3\\ (\%)\end{tabular}} & \multicolumn{1}{c|}{\begin{tabular}[c]{@{}c@{}}BLEU-4\\ (\%)\end{tabular}} & \multicolumn{1}{c|}{\begin{tabular}[c]{@{}c@{}}ROUGE-L\\ (\%)\end{tabular}} & \begin{tabular}[c]{@{}c@{}}METEOR\\ (\%)\end{tabular} & \multicolumn{1}{c|}{\begin{tabular}[c]{@{}c@{}}F1\\ (\%)\end{tabular}} & \begin{tabular}[c]{@{}c@{}}ACC\\ (\%)\end{tabular} \\ \hline
64 & \multicolumn{1}{c|}{68.00} & \multicolumn{1}{c|}{65.71} & \multicolumn{1}{c|}{87.37} & 90.63 & \multicolumn{1}{c|}{98.17} & 95.62 \\
128 & \multicolumn{1}{c|}{\textbf{74.27}} & \multicolumn{1}{c|}{\textbf{71.88}} & \multicolumn{1}{c|}{\textbf{89.87}} & \textbf{92.40} & \multicolumn{1}{c|}{\textbf{99.00}} & \textbf{96.25} \\
256 & \multicolumn{1}{c|}{71.80} & \multicolumn{1}{c|}{69.04} & \multicolumn{1}{c|}{87.88} & 90.23 & \multicolumn{1}{c|}{97.33} & 95.51 \\ \hline
\end{tabular}%
}
\end{table}

\section{Discussion and Conclusion}

We presented Surgical AI Copilot, a novel \textit{Planner–Worker} LLM agent framework for pituitary surgery that combines multi-step reasoning, tool invocation, and language-driven prompting through two core innovations: the PitAgent dataset, a richly annotated benchmark for surgical planning, and DEFT-GaLore, a Fourier-based low-rank adaptation method for efficient and effective LLM fine-tuning. Key findings highlight that this combination consistently improves the agent’s planning accuracy and prompt quality compared to existing PEFT baselines, demonstrating its effectiveness in surgical decision support. The Copilot effectively decomposes complex surgical queries, selects appropriate visual tools, and generates high-quality prompts, even in zero-shot settings, demonstrating its utility in real-time intraoperative workflows. However, the current system is limited to a single surgical domain. 
In this work, we validate the system only on a single visual model focused on surgical VQA. Future work will focus on expanding to other text-prompted visual models, such as segmentation, instrument detection, and overlaying, and evaluating the agent’s end-to-end performance across broader surgical scenarios, with the goal of developing a more generalizable, memory-augmented, and robust AI assistant for real-time surgical decision-making.

\section{Acknowledgements}
This work was supported by the EPSRC under grant [EP/W00805X/1, EP/Z534754/1]. HJM was supported by the NIHR UCLH/UCL Biomedical Research Centre. DZK was supported by Cleveland Clinic London for this work.  Thanks to Medtronic for access to the \textit{Touch Surgery\textsuperscript{TM} Ecosystem} for video recording, annotation, and storage. HJM is employed by and holds shares in Panda Surgical Ltd.

\bibliography{aaai2026}

@article{he2025pitvqa++,
  title={Pitvqa++: Vector matrix-low-rank adaptation for open-ended visual question answering in pituitary surgery},
  author={He, Runlong and Khan, Danyal Z and Mazomenos, Evangelos B and Marcus, Hani J and Stoyanov, Danail and Clarkson, Matthew J and Islam, Mobarakol},
  journal={arXiv preprint arXiv:2502.14149},
  year={2025}
}

@article{gao2024empowering,
  title={Empowering biomedical discovery with AI agents},
  author={Gao, Shanghua and Fang, Ada and Huang, Yepeng and Giunchiglia, Valentina and Noori, Ayush and Schwarz, Jonathan Richard and Ektefaie, Yasha and Kondic, Jovana and Zitnik, Marinka},
  journal={Cell},
  volume={187},
  number={22},
  pages={6125--6151},
  year={2024},
  publisher={Elsevier}
}

@article{maier2022surgical,
  title={Surgical data science--from concepts toward clinical translation},
  author={Maier-Hein, Lena and Eisenmann, Matthias and Sarikaya, Duygu and M{\"a}rz, Keno and Collins, Toby and Malpani, Anand and Fallert, Johannes and Feussner, Hubertus and Giannarou, Stamatia and Mascagni, Pietro and others},
  journal={Medical image analysis},
  volume={76},
  pages={102306},
  year={2022},
  publisher={Elsevier}
}

@article{chen2024vs,
  title={VS-assistant: versatile surgery assistant on the demand of surgeons},
  author={Chen, Zhen and Luo, Xingjian and Wu, Jinlin and Chan, Danny and Lei, Zhen and Wang, Jinqiao and Ourselin, Sebastien and Liu, Hongbin},
  journal={arXiv preprint arXiv:2405.08272},
  year={2024}
}

@article{khan2023current,
  title={Current and future advances in surgical therapy for pituitary adenoma},
  author={Khan, Danyal Z and Hanrahan, John G and Baldeweg, Stephanie E and Dorward, Neil L and Stoyanov, Danail and Marcus, Hani J},
  journal={Endocrine Reviews},
  volume={44},
  number={5},
  pages={947--959},
  year={2023},
  publisher={The Endocrine Society}
}

@inproceedings{gao2025multimodal,
title={Multi-modal Agent Tuning: Building a {VLM}-Driven Agent for Efficient Tool Usage},
author={Zhi Gao and Bofei Zhang and Pengxiang Li and Xiaojian Ma and Tao Yuan and Yue Fan and Yuwei Wu and Yunde Jia and Song-Chun Zhu and Qing Li},
booktitle={The Thirteenth International Conference on Learning Representations},
year={2025},
}

@article{kelly2024visiongpt,
  title={Visiongpt: Vision-language understanding agent using generalized multimodal framework},
  author={Kelly, Chris and Hu, Luhui and Yang, Bang and Tian, Yu and Yang, Deshun and Yang, Cindy and Huang, Zaoshan and Li, Zihao and Hu, Jiayin and Zou, Yuexian},
  journal={arXiv preprint arXiv:2403.09027},
  year={2024}
}

@article{tang2023medagents,
  title={Medagents: Large language models as collaborators for zero-shot medical reasoning},
  author={Tang, Xiangru and Zou, Anni and Zhang, Zhuosheng and Li, Ziming and Zhao, Yilun and Zhang, Xingyao and Cohan, Arman and Gerstein, Mark},
  journal={arXiv preprint arXiv:2311.10537},
  year={2023}
}

@inproceedings{wu2024surgbox,
  title={SurgBox: Agent-Driven Operating Room Sandbox with Surgery Copilot},
  author={Wu, Jinlin and Liang, Xusheng and Bai, Xuexue and Chen, Zhen},
  booktitle={2024 IEEE International Conference on Big Data (BigData)},
  pages={2041--2048},
  year={2024},
  organization={IEEE}
}

@article{gao2023retrieval,
  title={Retrieval-augmented generation for large language models: A survey},
  author={Gao, Yunfan and Xiong, Yun and Gao, Xinyu and Jia, Kangxiang and Pan, Jinliu and Bi, Yuxi and Dai, Yi and Sun, Jiawei and Wang, Haofen and Wang, Haofen},
  journal={arXiv preprint arXiv:2312.10997},
  volume={2},
  year={2023}
}

@article{hu2022lora,
  title={Lora: Low-rank adaptation of large language models.},
  author={Hu, Edward J and Shen, Yelong and Wallis, Phillip and Allen-Zhu, Zeyuan and Li, Yuanzhi and Wang, Shean and Wang, Lu and Chen, Weizhu and others},
  journal={ICLR},
  volume={1},
  number={2},
  pages={3},
  year={2022}
}

@inproceedings{liu2024dora,
  title={Dora: Weight-decomposed low-rank adaptation},
  author={Liu, Shih-Yang and Wang, Chien-Yi and Yin, Hongxu and Molchanov, Pavlo and Wang, Yu-Chiang Frank and Cheng, Kwang-Ting and Chen, Min-Hung},
  booktitle={Forty-first International Conference on Machine Learning},
  year={2024}
}

@article{zhao2024galore,
  title={Galore: Memory-efficient llm training by gradient low-rank projection},
  author={Zhao, Jiawei and Zhang, Zhenyu and Chen, Beidi and Wang, Zhangyang and Anandkumar, Anima and Tian, Yuandong},
  journal={arXiv preprint arXiv:2403.03507},
  year={2024}
}

@article{mao2024pitsurgrt,
title={PitSurgRT: real-time localization of critical anatomical structures in endoscopic pituitary surgery},
author={Mao, Zhehua and Das, Adrito and Islam, Mobarakol and Khan, Danyal Z and Williams, Simon C and Hanrahan, John G and Borg, Anouk and Dorward, Neil L and Clarkson, Matthew J and Stoyanov, Danail and Marcus, Hani J and Bano, Sophia},
journal={International Journal of Computer Assisted Radiology and Surgery},
year={2024}
}

@inproceedings{islam2020ap,
title={AP-MTL: Attention Pruned Multi-task Learning Model for Real-time Instrument Detection and Segmentation in Robot-assisted Surgery},
author={Islam, Mobarakol and VS, Vibashan and Ren, Hongliang},
booktitle={IEEE International Conference on Robotics and Automation (ICRA)},
year={2020}
}

@inproceedings{zhao2024transferring,
title={Transferring Knowledge from High-Quality to Low-Quality MRI for Adult Glioma Diagnosis},
author={Zhao, Yanguang and Bai, Long and Zhang, Zhaoxi and Wu, Yanan and Islam, Mobarakol and Ren, Hongliang},
booktitle={BraTS-SSA Challenge, International Conference on Medical Image Computing and Computer-Assisted Intervention (MICCAI) Workshop},
year={2024}
}

@article{enkaoua2023image,
title={Image-guidance in endoscopic pituitary surgery: an in-silico study of errors involved in tracker-based techniques},
author={Enkaoua, Aure and Islam, Mobarakol and Ramalhinho, João and Dowrick, Thomas and Booker, James and Khan, Danyal Z and Marcus, Hani J and Clarkson, Matthew J},
journal={Frontiers in Surgery},
year={2023}
}

@article{jiang2024mora,
title={MoRA: High-Rank Updating for Parameter-Efficient Fine-Tuning},
author={Jiang, Ting and Huang, Shaohan and Luo, Shengyue and Zhang, Zihan and Huang, Haizhen and Wei, Furu and Deng, Weiwei and Sun, Feng and Zhang, Qi and Wang, Songtao and Wang, Deqing and Zhuang, Fuzhen},
journal={arXiv preprint arXiv:2405.12130},
year={2024}
}

@inproceedings{moghani2024sufia,
  title={SuFIA: language-guided augmented dexterity for robotic surgical assistants},
  author={Moghani, Masoud and Doorenbos, Lars and Panitch, William Chung-Ho and Huver, Sean and Azizian, Mahdi and Goldberg, Ken and Garg, Animesh},
  booktitle={2024 IEEE/RSJ International Conference on Intelligent Robots and Systems (IROS)},
  pages={6969--6976},
  year={2024},
  organization={IEEE}
}

@article{llama32vision,
  title        = {The Llama 3 Herd of Models},
  author       = {Grattafiori, Aaron and Dubey, Abhimanyu and Jauhri, Abhinav and Pandey, Abhinav and Kadian, Abhishek and Al‑Dahle, Ahmad and … and many others},
  journal      = {arXiv preprint},
  year         = {2024},
  eid          = {2407.21783},
  url          = {https://arxiv.org/abs/2407.21783}
}

@inproceedings{he2024pitvqa,
  title={Pitvqa: Image-grounded text embedding llm for visual question answering in pituitary surgery},
  author={He, Runlong and Xu, Mengya and Das, Adrito and Khan, Danyal Z and Bano, Sophia and Marcus, Hani J and Stoyanov, Danail and Clarkson, Matthew J and Islam, Mobarakol},
  booktitle={International Conference on Medical Image Computing and Computer-Assisted Intervention},
  pages={488--498},
  year={2024},
  organization={Springer}
}

@article{zhu2025qwen,
  title={Qwen-2.5 outperforms other large language models in the chinese national nursing licensing examination: Retrospective cross-sectional comparative study},
  author={Zhu, Shiben and Hu, Wanqin and Yang, Zhi and Yan, Jiani and Zhang, Fang},
  journal={JMIR Medical Informatics},
  volume={13},
  pages={e63731},
  year={2025},
  publisher={JMIR Publications Toronto, Canada}
}

@article{modoranu2025svd,
  title={SVD-Free Low-Rank Adaptive Gradient Optimization for Large Language Models},
  author={Modoranu, Ionut-Vlad and Safaryan, Mher and Schultheis, Erik and Alistarh, Dan},
  journal={arXiv preprint arXiv:2505.17967},
  year={2025}
}

@article{dutta2024adaptive,
  title={Adaptive reasoning and acting in medical language agents},
  author={Dutta, Abhishek and Hsiao, Yen-Che},
  journal={arXiv preprint arXiv:2410.10020},
  year={2024}
}

@article{wei2024medaide,
  title={Medaide: Towards an omni medical aide via specialized llm-based multi-agent collaboration},
  author={Wei, Jinjie and Yang, Dingkang and Li, Yanshu and Xu, Qingyao and Chen, Zhaoyu and Li, Mingcheng and Jiang, Yue and Hou, Xiaolu and Zhang, Lihua},
  journal={arXiv preprint arXiv:2410.12532},
  year={2024}
}

@article{su2025galore,
  title={Galore 2: Large-scale llm pre-training by gradient low-rank projection},
  author={Su, DiJia and Gu, Andrew and Xu, Jane and Tian, Yuandong and Zhao, Jiawei},
  journal={arXiv preprint arXiv:2504.20437},
  year={2025}
}

@article{team2024gemma,
  title={Gemma 2: Improving open language models at a practical size},
  author={Team, Gemma and Riviere, Morgane and Pathak, Shreya and Sessa, Pier Giuseppe and Hardin, Cassidy and Bhupatiraju, Surya and Hussenot, L{\'e}onard and Mesnard, Thomas and Shahriari, Bobak and Ram{\'e}, Alexandre and others},
  journal={arXiv preprint arXiv:2408.00118},
  year={2024}
}

@article{bi2024deepseek,
  title={Deepseek llm: Scaling open-source language models with longtermism},
  author={Bi, Xiao and Chen, Deli and Chen, Guanting and Chen, Shanhuang and Dai, Damai and Deng, Chengqi and Ding, Honghui and Dong, Kai and Du, Qiushi and Fu, Zhe and others},
  journal={arXiv preprint arXiv:2401.02954},
  year={2024}
}

@article{zhang2025parameter,
  title={Parameter-efficient fine-tuning for foundation models},
  author={Zhang, Dan and Feng, Tao and Xue, Lilong and Wang, Yuandong and Dong, Yuxiao and Tang, Jie},
  journal={arXiv preprint arXiv:2501.13787},
  year={2025}
}

@article{han2024parameter,
  title={Parameter-efficient fine-tuning for large models: A comprehensive survey},
  author={Han, Zeyu and Gao, Chao and Liu, Jinyang and Zhang, Jeff and Zhang, Sai Qian},
  journal={arXiv preprint arXiv:2403.14608},
  year={2024}
}

@article{chen2023parameter,
  title={Parameter-efficient fine-tuning design spaces},
  author={Chen, Jiaao and Zhang, Aston and Shi, Xingjian and Li, Mu and Smola, Alex and Yang, Diyi},
  journal={arXiv preprint arXiv:2301.01821},
  year={2023}
}

@article{reza2019class,
  title={A class of randomized Subset Selection Methods for large complex networks},
  author={Reza, Amit and Tripathi, Richa},
  journal={arXiv preprint arXiv:1905.04452},
  year={2019}
}

@article{fein2025fft,
  title={The FFT Strikes Back: An Efficient Alternative to Self-Attention},
  author={Fein-Ashley, Jacob},
  journal={arXiv e-prints},
  pages={arXiv--2502},
  year={2025}
}

@article{xu2023rewoo,
  title={Rewoo: Decoupling reasoning from observations for efficient augmented language models},
  author={Xu, Binfeng and Peng, Zhiyuan and Lei, Bowen and Mukherjee, Subhabrata and Liu, Yuchen and Xu, Dongkuan},
  journal={arXiv preprint arXiv:2305.18323},
  year={2023}
}

@inproceedings{zeng2024agenttuning,
  title={AgentTuning: Enabling Generalized Agent Abilities for LLMs},
  author={Zeng, Aohan and Liu, Mingdao and Lu, Rui and Wang, Bowen and Liu, Xiao and Dong, Yuxiao and Tang, Jie},
  booktitle={Findings of the Association for Computational Linguistics ACL 2024},
  pages={3053--3077},
  year={2024}
}

@inproceedings{yang2018hotpotqa,
  title={HotpotQA: A Dataset for Diverse, Explainable Multi-hop Question Answering},
  author={Yang, Zhilin and Qi, Peng and Zhang, Saizheng and Bengio, Yoshua and Cohen, William and Salakhutdinov, Ruslan and Manning, Christopher D},
  booktitle={Proceedings of the 2018 Conference on Empirical Methods in Natural Language Processing},
  pages={2369--2380},
  year={2018}
}

@inproceedings{joshi2017triviaqa,
  title={TriviaQA: A Large Scale Distantly Supervised Challenge Dataset for Reading Comprehension},
  author={Joshi, Mandar and Choi, Eunsol and Weld, Daniel S and Zettlemoyer, Luke},
  booktitle={Proceedings of the 55th Annual Meeting of the Association for Computational Linguistics (Volume 1: Long Papers)},
  pages={1601--1611},
  year={2017}
}
\newpage
\appendix
\section{Appendix A: System Message for the Agent Planning}
A system message is an instruction given to a LLM to guide its behavior, tone, or role throughout the conversation. In our work, the system message specifies the role of LLM, the overall task, the available text-promptable downstream models and examples of response.  We tested 20 different system messages on the zero-shot LLaMA 3.2 and Qwen 2.5 models to evaluate their optimal task planning and prompt generation performance. After testing, the system message we selected is as follows:

\begin{tcolorbox}[
    colback=gray!15,
    colframe=white,
    boxrule=0pt,
    arc=0pt,
    boxsep=5pt,
    left=8pt,
    right=8pt,
    before upper={\parindent 0pt}
]
You are a surgical AI agent assisting in pituitary surgery. Your job is to handle surgeons' queries efficiently by choosing appropriate text-promptable AI models and generating corresponding prompts.\\
Available models: Segment-Video, Segment-MRI, Track-Instrument, Surgical-VQA, Overlaying.\\
- Use one model if query focuses on a single, simple aspect:\\
Example (single-model):\\

\begin{tcolorbox}[
    colback=gray!25,
    colframe=white,
    boxrule=0pt,
    arc=0pt,
    left=8pt,
    top=4pt,
    bottom=4pt,
    width=\linewidth
]
Model: Segment-Video\\Prompt: Segment the sella in the video.
\end{tcolorbox}

- Use MULTIPLE models if query requires several types of information:\\
Example (multi-model):\\

\begin{tcolorbox}[
    colback=gray!25,
    colframe=white,
    boxrule=0pt,
    arc=0pt,
    left=8pt,
    top=4pt,
    bottom=4pt,
    width=\linewidth
]
Step1:\\Model: Segment-MRI\\Prompt: Segment the pituitary tumor from MRI.\\
Step2:\\Model: Segment-Video\\Prompt: Segment the sella in the video.
\end{tcolorbox}

Now, follow the same format to answer the provided question-no extra text, labels, or formatting.
\end{tcolorbox}

\section*{Appendix B: Algorithm for DEFT-GaLore}

DEFT-GaLore is a memory-efficient training method that significantly reduces optimizer state memory. Instead of using costly SVD, it employs a highly efficient energy-based FFT projection to identify the gradient's most important directions. 
This projection is updated periodically to amortize computational cost.
The full gradient is projected into a low-rank subspace, where the compact AdamW optimizer states (moments) are maintained and updated. This low-rank update is then projected back to the original dimension to update the model's weights, followed by a standard decoupled weight decay step. This approach effectively balances substantial memory savings with high computational performance.

\begin{algorithm}[H] 
\caption{AdamW with DEFT-GaLore}
\label{alg:adamw-galore}
\begin{algorithmic}[1] 
    \Require Layer weights $\mathbf{W} \in \mathbb{R}^{m \times n}$, learning rate $\eta$, betas $\beta_1, \beta_2$, epsilon $\epsilon$, weight decay $\lambda$, rank $k$, subspace update frequency $T$.
    \State Initialize $\mathbf{M}_0 \leftarrow 0$, $\mathbf{V}_0 \leftarrow 0$, step $t \leftarrow 0$.
    \State Initialize projection matrix $P_t \leftarrow \text{None}$.
    \Repeat
        \State $G_t \leftarrow \nabla_{\mathbf{W}_t}\mathcal{L}(\mathbf{W}_t)$ \Comment{Compute gradient}
        \If{$t \pmod{T} = 0$ \textbf{or} $P_t$ is None}
            \State $P_t \leftarrow \text{DEFT-Projection}(G_t, r)$ {Update projector (see Alg. 2)}
        \Else
            \State $P_t \leftarrow P_{t-1}$ \Comment{Reuse previous projector}
        \EndIf
        \State $\tilde{G_t} \leftarrow P_t^T G_t$ {Project gradient into low-rank space}
        \State $t \leftarrow t + 1$
        \State $\mathbf{M}_t \leftarrow \beta_1 \cdot \mathbf{M}_{t-1} + (1 - \beta_1) \cdot \tilde{G_t}$
        \State $\mathbf{V}_t \leftarrow \beta_2 \cdot \mathbf{V}_{t-1} + (1 - \beta_2) \cdot \tilde{G_t}^2$
        \State $\hat{\mathbf{M}}_t \leftarrow \mathbf{M}_t / (1 - \beta_1^t)$ \Comment{Bias correction}
        \State $\hat{\mathbf{V}}_t \leftarrow \mathbf{V}_t / (1 - \beta_2^t)$
        \State $N_t \leftarrow \hat{\mathbf{M}}_t / (\sqrt{\hat{\mathbf{V}}_t} + \epsilon)$ \Comment{Compute low-rank update}
        \State $\tilde{G}_t \leftarrow P_t N_t$ \Comment{Project update back to original space}
        \State $\mathbf{W}_t \leftarrow \mathbf{W}_{t-1} - \eta \cdot \tilde{G}_t$ \Comment{Update weights}
        \State $\mathbf{W}_t \leftarrow \mathbf{W}_t - \eta \cdot \lambda \cdot \mathbf{W}_t$ \Comment{Apply decoupled weight decay}
    \Until{convergence criteria met}
    \State \textbf{return} $\mathbf{W}_T$ 

\end{algorithmic}
\end{algorithm}

\begin{algorithm}[H]
\caption{{DEFT-Projection}}
\label{alg:fft_proj}
\begin{algorithmic}[1]
\Require Matrix $G_t \in \mathbb{R}^{m \times n}$, target rank $k$.
\Ensure Orthogonal basis $M \in \mathbb{R}^{m \times k}$.
\State $G_{t,f} \leftarrow \text{FFT}(G_t)$ \Comment{Apply FFT along columns}
\State $\text{score} \leftarrow \sum_{i=0}^{m-1} |G_{t,f}|^2$ \Comment{Calculate energy score per frequency}
\State $\text{idx} \leftarrow$ indices of top $k$ values in score
\State $Y_c \leftarrow G_{t,f}[:, \text{idx}]$ \Comment{Sketch with top-k complex components}
\State $Q_c, \_ \leftarrow \text{QR}(Y_c)$ \Comment{Complex QR decomposition}
\State $Y_r \leftarrow \text{Reshape}(\text{RealImag}(Q_c))$ to matrix in $\mathbb{R}^{m \times 2k}$
\State $Q_r, \_ \leftarrow \text{QR}(Y_r)$ \Comment{Real QR decomposition}
\State \textbf{return} $Q_r[:, :k]$ \Comment{Return first k orthogonal vectors}
\end{algorithmic}
\end{algorithm}

In the main text, we named the projection matrix $M_t$. Here, to distinguish it from the AdamW momentum, we name the projection matrix $P_t$.

\section*{Appendix C: Supplementary Implementation Details of The \textit{Planner} Fine-tuning Experiment}
Fo the PEFT methods based on learning parameters (LoRA~\cite{hu2022lora}, DoRA~\cite{liu2024dora}, MoRA~\cite{jiang2024mora}), We set their learning rate (LR) to $3 \times 10^{-7}$, batch size to 6, number of epochs to 10, rank to 8, dropout to 0.3, and scaling factor to 16.

For PEFT methods based on gradient low-rank projection (GaLore~\cite{zhao2024galore}, GaLore 2~\cite{su2025galore}, DCT-GaLore~\cite{modoranu2025svd}, and our DEFT-GaLore), we also set their learning rate to $3 \times 10^{-7}$, batch size to 6, number of epochs to 5, rank to 128, projection matrix update interval to 50, scaling factor to 1, and projection strategy to \textit{reverse\_std}. All other hyperparameters use the default settings of Galore\footnote{https://github.com/jiaweizzhao/GaLore}.

In the GaLore design, matrix projection has multiple modes. The standard mode (\textit{std}) adapts the projection direction based on the shape of the gradient matrix. When the height of the matrix is greater than or equal to the width, right projection $G_t Q_t$ is performed, and otherwise, left projection $P_t^T G_t$ is performed. This strategy selects the smaller of the matrix dimensions for projection. The reverse std method performs the inverse operation and left projection  when the height of the matrix is greater than or equal to the width. This strategy selects the larger of the matrix dimensions for projection. We use the same projection design when reproducing the DCT-GaLore and designing the DEFT-GaLore method.

\section*{Appendix D: Validation on Rewoo Planner Benchmark}
We also validate our DEFT-GaLore on a benchmark agent planning dataset: the ReWOO Planner Instruction Tuning dataset. This dataset, released by the ReWOO team~\footnote{\url{https://huggingface.co/datasets/rewoo/planner_instruction_tuning_2k}} and referred to as ``rewoo/planner\_instruction\_tuning\_2k" on Hugging Face~\cite{xu2023rewoo}, contains 2,000 high-quality examples designed to fine-tune the planning component of large language models. It emphasizes explicit multi-step reasoning by providing planning trajectories for complex question-answering tasks, particularly from HotpotQA~\cite{yang2018hotpotqa} and TriviaQA~\cite{joshi2017triviaqa}. Each example includes an instruction, an input query, and a sequence of structured planning steps that incorporate external tool calls (e.g., Wikipedia[input]) and reference intermediate evidence variables (e.g., \#E1, \#E2). This setup allows models to learn how to decompose a task into executable steps and gather supporting evidence before generating a final answer. In our experiments, we used 1,500 samples for training and held out the remaining 500 for testing, following the setup in~\cite{zeng2024agenttuning}. This dataset serves as a valuable resource for evaluating and improving the planning and reasoning capabilities of ReWOO-style agents.

To implement this experiment, we keep the same configuration as the Planner fine-tuning experiment except for changing learning rate, seed, and number of epochs to $5 \times 10^{-5}$, 2025, and 15 respectively.

\begin{table}[H] 
\centering
\caption{Comparison of PEFT methods on the ReWOO Planner Instruction Tuning dataset for multi-step planning performance. Our method, DEFT-GaLore, achieves the best results across BLEU-4, ROUGE-L, and METEOR, demonstrating state-of-the-art performance in structured multi-step planning for LLM-based agents.}
\label{tab:ReWOO}
\scalebox{0.75}{
\begin{tabular}{c|c|c|c}
\hline
Method & \begin{tabular}[c]{@{}c@{}}BLEU-4\\ (\%)\end{tabular} & \begin{tabular}[c]{@{}c@{}}ROUGE-L\\ (\%)\end{tabular} & \begin{tabular}[c]{@{}c@{}}METEOR\\ (\%)\end{tabular} \\ \hline
Zero-Shot  & 2.1 & 15.3 & 21.4 \\ \hline
LoRA~\cite{hu2022lora} & 52.5 & 68.7 & 76.1 \\ \hline
DoRA~\cite{liu2024dora}  & 53.5 & 69.2 & 76.1 \\ \hline
MoRA~\cite{jiang2024mora}  & 50.9 & 67.6 & 75.7 \\ \hline
GaLore~\cite{zhao2024galore} & 58.5 & 71.8 & 78.7 \\ \hline
GaLore 2~\cite{su2025galore} & 62.3 & 73.9 & 80.1 \\ \hline
DCT-GaLore~\cite{modoranu2025svd} & 62.7 & 74.2 & 80.7 \\ \hline
DEFT-GaLore (ours) & \textbf{64.9} & \textbf{75.5} & \textbf{81.8} \\ \hline
\end{tabular}
}
\end{table}

Table 1 showcases the multi-step planning performance of our method, DEFT-GaLore, compared to other state-of-the-art PEFT approaches on the benchmark ReWOO Planner Instruction Tuning dataset. DEFT-GaLore consistently outperforms prior methods, achieving up to 3.5\% absolute improvement in BLEU-4, 1.3\% in ROUGE-L, and 1.1\% in METEOR over the best previous SOTA (DCT-GaLore). These gains demonstrate the effectiveness of our FFT-based low-rank adaptation in enhancing the planning capability of LLM agents for structured multi-step reasoning tasks.

\end{document}